\documentclass{article}

\PassOptionsToPackage{numbers, compress}{natbib}

\usepackage[sglblindworkshop, final]{neurips_2025}

\usepackage[utf8]{inputenc} 
\usepackage[T1]{fontenc}    
\usepackage{hyperref}       
\usepackage{url}            
\usepackage{booktabs}       
\usepackage{amsfonts}       
\usepackage{nicefrac}       
\usepackage{microtype}      
\usepackage{xcolor}         
\usepackage{graphicx}
\usepackage{amsmath}
\title{Diffusion-Augmented Reinforcement Learning for Robust Portfolio Optimization under Stress Scenarios}
\workshoptitle{GenAI in Finance}

\vspace{-7mm}

\author{
  \fontsize{12pt}{14pt}\selectfont
  \href{mailto:d21024@students.iitmandi.ac.in}{\textbf{Himanshu Choudhary*}} \\[0.5em]
  \href{mailto:d21022@students.iitmandi.ac.in}{\textbf{Arishi Orra}} \\[0.5em]
  \href{mailto:manoj@iitmandi.ac.in}{\textbf{Manoj Thakur}} \\[0.5em]
  Indian Institute of Technology Mandi, India, 175005. \\
  Mail ID*: d21024@students.iitmandi.ac.in
}

\begin{document}

\maketitle

\vspace{-7mm}

\begin{abstract}
  In the ever-changing and intricate landscape of financial markets, portfolio optimisation remains a formidable challenge for investors and asset managers. Conventional methods often struggle to capture the complex dynamics of market behaviour and align with diverse investor preferences. To address this, we propose an innovative framework, termed Diffusion-Augmented Reinforcement Learning (DARL), which synergistically integrates Denoising Diffusion Probabilistic Models (DDPMs) with Deep Reinforcement Learning (DRL) for portfolio management. By leveraging DDPMs to generate synthetic market crash scenarios conditioned on varying stress intensities, our approach significantly enhances the robustness of training data. Empirical evaluations demonstrate that DARL outperforms traditional baselines, delivering superior risk-adjusted returns and resilience against unforeseen crises, such as the $2025$ Tariff Crisis. This work offers a robust and practical methodology to bolster stress resilience in DRL-driven financial applications.
\end{abstract}


\section{Introduction}
    Portfolio Optimisation has long been regarded as one of the central problems in financial decision-making. The objective of portfolio optimization is to allocate capital among a set of assets in order to balance risk and return. The foundations of modern portfolio theory can be traced back to the pioneering work of Markowitz \cite{markowitz1952}, who first introduced the mean–variance framework. This contribution laid the groundwork for much of the subsequent development in financial economics such as \citep{Levy1970, Davis1990, Anagnostopoulos2010}. However, these classical methods rely on strong assumptions such as normally distributed returns and stationarity of data. In practice, financial markets are noisy, non-stationary, and often exhibit fat-tailed behaviour. Moreover, such models are typically static in nature and do not adapt dynamically to changing market regimes, which limits their usefulness in real-world trading environments.
    
    Reinforcement Learning (RL) has emerged as a dynamic alternative, treating portfolio allocation as a Markov Decision Process (MDP) where agents learn policies from market states \cite{Sutton1998}. RL offers a natural sequential decision-making framework, where the portfolio allocation problem is cast as an agent interacting with a financial environment. The agent receives rewards based on the performance of its allocations and learns a policy that maximises long-term cumulative return. By continuously updating its policy, RL models can adapt to changing market conditions, learn complex non-linear relationships, and dynamic rebalancing strategies \citep{Jiang2017, Liu2022, Choudhary2025, ChouFinXplore, OrraExpert2025}. However, these models trained on historical data often underperform in crises due to distribution shifts. Parallel to the advances in RL, generative models have gained significant traction in finance, particularly for scenario generation and stress testing \cite{Flaig2022, Zheng2024}. Early efforts involved using bootstrapping and Monte Carlo techniques for simulating return distributions. Later, with the advent of deep generative models such as Variational Autoencoders (VAEs) and Generative Adversarial Networks (GANs), researchers were able to generate more realistic financial time series that capture the heavy tails, volatility clustering, and regime shifts inherent in market data \cite{Wiese2020}. These synthetic scenarios are useful both for portfolio stress testing and for augmenting limited training data in machine learning-based strategies. Data augmentation techniques, common in computer vision, are underexplored in finance due to the sequential and correlated nature of time-series data. Diffusion models, particularly Denoising Diffusion Probabilistic Models (DDPMs) \cite{Ho2020}, excel at generating realistic samples by reversing a noise-addition process and can be conditioned for targeted scenarios like market crashes. The integration of diffusion models with RL thus offers a twofold benefit: (i) it mitigates the data scarcity and non-stationarity problem in financial markets by providing diverse synthetic scenarios, and (ii) it strengthens the ability of RL agents to learn stable and risk-aware portfolio policies. This hybrid paradigm represents a significant step forward in bridging the gap between theoretical models and practical investment strategies. In this study, we introduce Diffusion-Augmented RL (DARL) for portfolio management. A DDPM-based generator for synthetic return sequences conditioned on crash intensity, augmenting training data with crisis-like scenarios. A custom RL environment using change in portfolio values as reward and covariance-inclusive states. Empirical validation on Dow $30$ stocks, showing improved robustness across crises, including the unseen $2025$ Tariff Crisis.

    \section{Problem Formulation}

    Portfolio Optimisation is essentially concerned with the systematic allocation of wealth across a set of financial assets in such a manner that the overall return is maximised while the associated risks are contained. Since financial markets are inherently stochastic and evolve over time, the problem is naturally suited to be modelled in the framework of a MDP. Formally, the portfolio optimisation task can be described as a tuple $(\mathcal{S}, \mathcal{A}, \mathbb{P}, \mathcal{R}, \gamma)$, where: $\mathcal{S}$ denotes the set of states, $\mathcal{A}$ denotes the set of possible actions, $\mathbb{P}$ represents the state transition probability distribution, $\mathcal{R}$ defines the reward function, and $\gamma \in [0,1]$ is the discount factor capturing the present value of future rewards. At any given time step $t$, the state $s_{t} \in \mathcal{S}$ contains information about the financial environment. In our setting, this includes historical and current stock prices, risk-related measures such as the covariance matrix of asset returns, and a selected set of technical indicators that summarise market trends. The action $w_{i} \in \mathcal{A}$ is represented by an $n$-dimensional vector, where each element specifies the proportion of the portfolio allocated to one of the $n$ assets under consideration. Naturally, the action space is constrained by $\sum_{i=1}^{n} w_{i} =1$ and $w_{i} \geq 0$, ensuring that the entire wealth is distributed among the available assets without short-selling. The reward $r_t \in \mathcal{R}$ at each time step is defined as the portfolio return realised after the execution of the chosen allocation. The primary goal of the agent is to learn an optimal policy $\pi^{*}$.

    \section{Proposed Methodology}

        Our DARL framework integrates DDPMs with Proximal Policy Optimization (PPO) \cite{Schulman2017} for robust portfolio management. The approach consists of diffusion-based data augmentation for crisis scenarios, diffusion-based regularization for portfolio weights, a custom RL environment, and an iterative training procedure. DRL methods, such as PPO, excel in sequential decision-making tasks like portfolio optimization by learning policies that map market states to allocation actions, maximizing cumulative rewards. However, standard DRL models trained on historical financial data often suffer from overfitting to stable market regimes, leading to poor generalization during rare, high-volatility events like financial crises. This is exacerbated by the non-stationary nature of financial time series, where distribution shifts (e.g., sudden return correlations or volatility spikes) violate the i.i.d. assumptions implicit in many RL algorithms. To address these limitations, we employ diffusion-based data augmentation, leveraging DDPMs to generate synthetic training data. Diffusion models can generate diverse, realistic samples conditioned on specific attributes (e.g., crash intensity), simulating tail-risk scenarios absent or underrepresented in historical data. Augmentation expands the training dataset without collecting new real data, mitigating data scarcity in finance where crises occur infrequently. This leads to better policy generalization, by encouraging the agent to learn invariant features across augmented and real data. Augmentation integrates seamlessly into DRL by appending synthetic episodes to the replay buffer or environment, enabling the agent to explore high-reward policies in simulated stress. Overall, diffusion augmentation bridges the gap between limited historical data and the need for crisis-resilient models, making DRL more viable for real-world portfolio management where black-swan events can erode years of gains.

        \subsection{Diffusion Model for Scenario Generation}

            We employ a \textit{Denoising Diffusion Probabilistic Model (DDPM)} to generate stress scenarios. 
            The forward diffusion process gradually perturbs clean data $x_0$ (asset return sequences) 
            with Gaussian noise over $T$ steps: $q(x_t \mid x_{t-1}) = \mathcal{N}\!\big(x_t; \sqrt{\alpha_t}\,x_{t-1}, \, \beta_t \mathbf{I}\big), q(x_t \mid x_0) = \mathcal{N}\!\big(x_t; \sqrt{\bar\alpha_t}\,x_0, \, (1-\bar\alpha_t)\mathbf{I}\big)$.
            where $\alpha_t = 1-\beta_t$, $\bar\alpha_t = \prod_{s=1}^t \alpha_s$, and $\beta_t$ follows a linear 
            schedule $[10^{-4},0.02]$. The reverse process is learned with a neural network that approximates: $p_\theta(x_{t-1}\mid x_t,c) = \mathcal{N}\!\Big(x_{t-1}; \, \mu_\theta(x_t,t,c), \, \tilde\beta_t \mathbf{I}\Big)$, where $c$ is a conditioning variable (e.g., crash intensity).
            We parameterize the mean via noise prediction as:
            \begin{equation*}
                \mu_\theta(x_t,t,c) = \frac{1}{\sqrt{\alpha_t}}\Bigg(x_t - \frac{\beta_t}{\sqrt{1-\bar\alpha_t}}\,\epsilon_\theta(x_t,t,c)\Bigg),
            \end{equation*}
            with the training objective given by the simplified mean-squared error loss: $\mathcal{L}(\theta) = \mathbb{E}_{x_0,\,\epsilon,\,t,\,c}\Big[ \, \|\epsilon - \epsilon_\theta(x_t,t,c)\|^2 \, \Big]$. At generation time, we start from $x_T \sim \mathcal{N}(0,I)$ and iteratively apply the reverse 
            process to sample synthetic market scenarios.

            Crash intensity helps our system create fake market crashes to train the PPO agent better. This makes the agent ready for tough market times, even with little real crash data. It learns from different fake crashes, helping it handle surprises like the $2025$ Tariff Crisis, using patterns from $2007-2009$ (Financial Crisis), $2020-2021$ (COVID-$19$), and $2025$. The final model works on real trade/test data without fake crashes but uses this training to stay strong and adjust well, with better results.

            \begin{figure}[!h]
        	\centering
        	\includegraphics[scale=0.4]{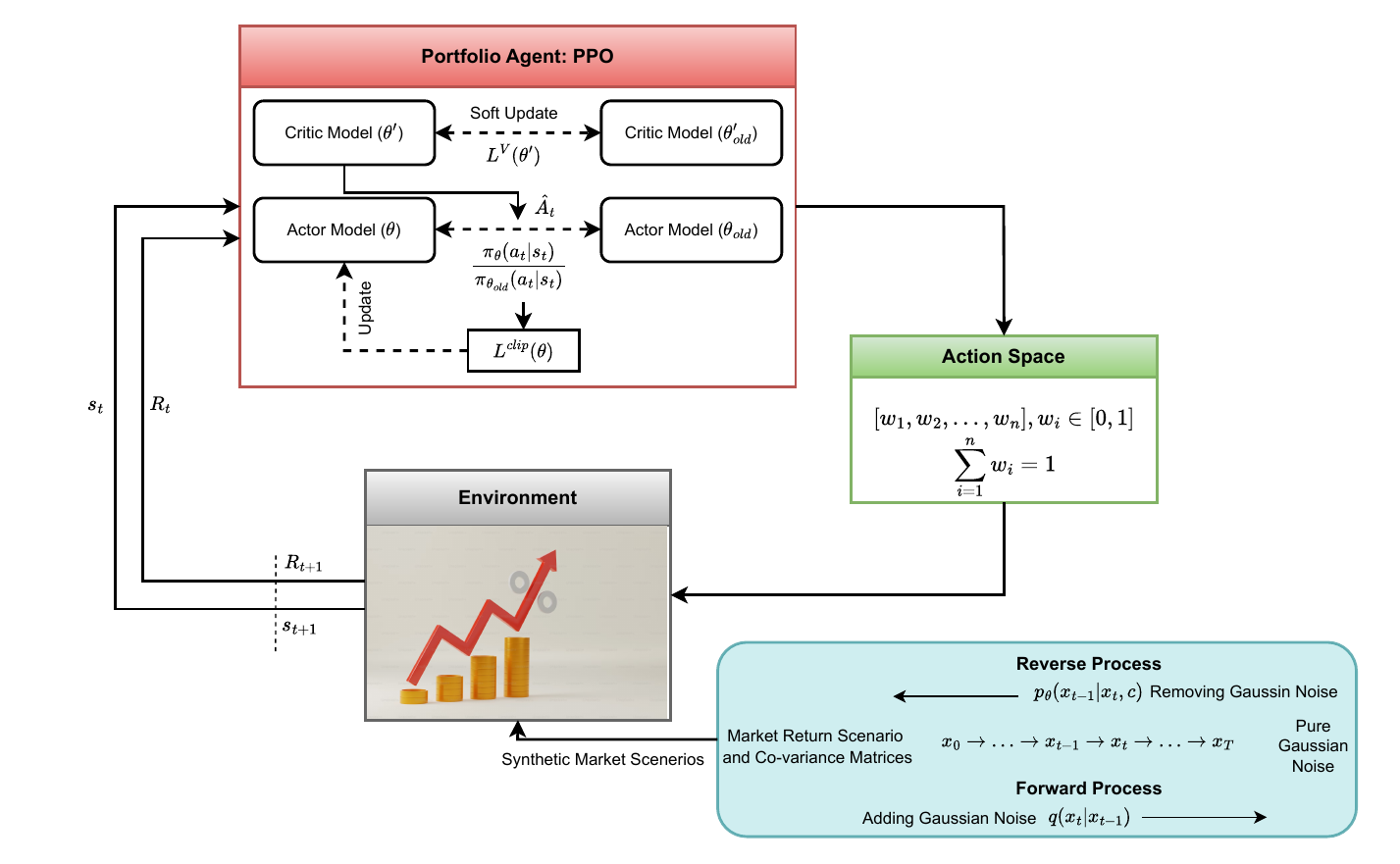}
        	\caption{The Proposed DARL Framework}
        	\label{Proposed}
        \end{figure}

    
    \section{Experiment}

    \subsection{Data Description and Experimental Settings}
        We used data from the Dow Jones Industrial Average (DJI), comprising its $30$ constituent stocks. Daily closing prices for all DJI stocks were retrieved from Yahoo Finance, covering $1$ January $2011$ to $31$ July $2025$. For model development, the training window spans $1$ January $2011$ to $31$ December $2023$, while the period $1$ January $2024$ to $31$ July $2025$ is held aside for out-of-sample evaluation. To reflect realistic trading conditions, an initial capital of $1$ million is provided to the agent for investment in the selected stocks. In line with the work of \cite{Choudhary2025, OrraDynamic2024}, a transaction cost of $0.05\%$ is charged on every trade to mirror market frictions. Hyperparameters for the deep reinforcement learning models are tuned via Bayesian optimization, to ensure well-calibrated policies. The search ranges for the hyperparameters were determined empirically to balance thorough exploration with computational practicality. Our environment is motivated by \citep{Liu2022, Choudhary2025, OrraICML2025}.

    \vspace*{-2mm}
    
    \subsection{Results and Discussion}
\begin{figure}[!h]
        	\centering
        	\includegraphics[scale=0.25]{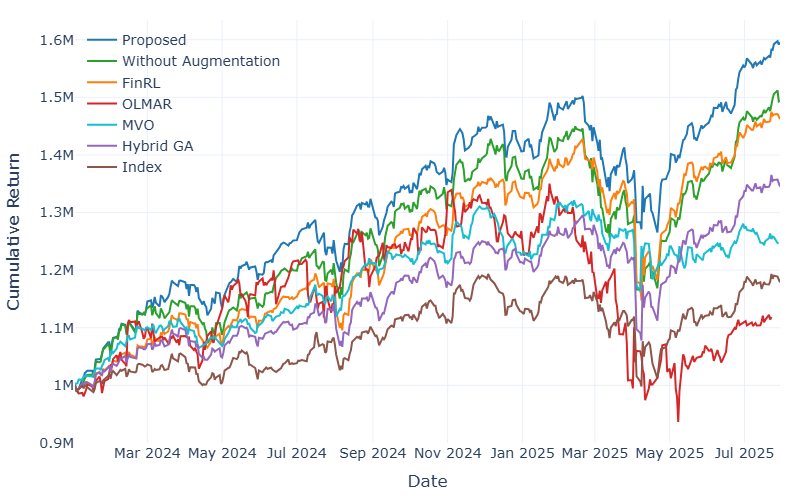}
        	\caption{Cumulative return trajectories of the proposed DARL and baseline models.}
        	\label{Cum}
        \end{figure}

        The performance of the proposed framework was benchmarked against several portfolio optimization models, including Without Augmentation, FinRL-PPO, Online Moving Average Reversion (OLMAR), Hybrid GA, Markowitz Model, and market index. Table \ref{Tab1} summarizes the performance indicators, while Figure \ref{Cum} illustrates the cumulative return trends over the entire trading period.

        The Proposed approach outperformed all benchmarks across multiple performance metrics. It achieved the highest cumulative return $(59.52\%)$ and annualized return $(34.71\%)$, substantially surpassing the market index ($17.88\%$ cumulative, $11.07\%$ annualized). This indicates that the proposed model consistently generated superior long-term returns compared to both traditional and AI-driven baselines. In terms of risk-adjusted performance, DARL demonstrated a Sharpe ratio of $1.91$ and a Calmar ratio of $2.20$, which are significantly higher than those of competing models. These results highlight its ability to provide superior returns while effectively controlling downside risk. When compared with the Without Augmentation setup, the proposed method showed considerable improvements across all metrics, underscoring the importance of data augmentation in enhancing predictive robustness and portfolio allocation decisions. Although Hybrid GA and FinRL-PPO exhibited competitive performance, they lagged behind the proposed framework in terms of drawdown management. Notably, the proposed model recorded a maximum drawdown of $-15.76\%$, which is relatively lower than Hybrid GA $(-16.79\%)$ and Without Augmentation $(-20.30\%)$, suggesting stronger resilience during market downturns. The cumulative wealth trajectories shown in Figure \ref{Cum} further validate these findings. The proposed method maintains a consistently upward trend, outperforming benchmarks throughout the trading horizon. During volatile market phases (Tariff Announcement), the proposed displayed faster recovery and more stable growth, whereas traditional approaches like OLMAR, Hybrid GA, and the index suffered sharp declines and slower recoveries. In summary, the results confirm that the proposed framework not only maximizes returns but also enhances risk-adjusted performance and portfolio stability, making it more effective for real-world trading environments.
        
        \vspace*{-3mm}
        \begin{table*}[!htp]
    		\centering
    		\caption{Performance indicators of the Proposed DARL approach and Benchmarks for entire trading period} \label{Tab1}
    		\resizebox{\textwidth}{!}{%
    			\begin{tabular}{lccccccccccc}\toprule
    				\textbf{Model/Benchmark} & \textbf{Cumulative Return (\%)} & \textbf{Annualized Return (\%)} & \textbf{Sharpe Ratio} & \textbf{Calmar Ratio} & \textbf{Annual Volatility (\%)} & \textbf{Maximum Drawdown (\%)} \\ \midrule
    				\textbf{Proposed}        & \textbf{59.5253}  & \textbf{34.7101}   & \textbf{1.9096}   & \textbf{2.2024}  & 16.3058  & -15.7598  \\
    				Without Augmentation     & 49.4439           & 29.2149            & 1.5172          & 1.4385           & 17.9649    & -20.3080          \\
    			  FinRL-PPO                & 46.2286           & 27.4344            & 1.5411          & 1.3961         & 16.6335    & -19.6496          \\
    				OLMAR                    & 11.8773           & 7.5214             & 0.4097            & 0.2455           & 25.6876    & -30.6370          \\
                    Hybrid-GA                & 34.5056           & 20.8184            & 1.2623            & 1.2403           & 15.9922    & -16.7852                       \\
                    Markowitz                & 24.6485           & 15.1333            & 1.1246            & 1.2754           & \textbf{13.3178}    & \textbf{-11.8651}                       \\
    				Index                   & 17.8874            & 11.0694            & 0.7717            & 0.6762           & 15.0674    & -16.3692     \\
    				\bottomrule
    			\end{tabular}
    		}
    	\end{table*}

        \vspace*{-3mm}

    \vspace*{-3mm}
    \section{Conclusion}
        The Diffusion-Augmented Reinforcement Learning (DARL) framework represents a significant advancement in addressing the challenges of portfolio optimisation in volatile financial markets. By integrating DDPMs with DRL, our approach effectively enhances the robustness of portfolio management strategies, particularly during the unseen 2025 Tariff Crisis. The use of conditional DDPMs to generate synthetic crash scenarios, parameterised by stress intensity, enriches the training data, enabling the PPO agent to learn policies resilient to tail-risk events. Empirical results on Dow $30$ stocks demonstrate DARL’s superior performance compared to traditional baselines. This methodology offers a practical and scalable solution for investors and asset managers, equipping them to navigate complex market dynamics with greater confidence. 
        
        Future research will focus on enhancing the DARL framework by integrating sentiment analysis derived from news articles and insights gleaned from company financial reports. This addition would equip the model with valuable context regarding market conditions, thereby enabling more effective responses to unforeseen events.



\end{document}